\documentclass[10pt,twocolumn,letterpaper]{article}
\usepackage[margin=2.5cm]{geometry}
\usepackage{times,epsfig,graphicx,amsmath,amssymb}
\usepackage[ruled,vlined]{algorithm2e}
\usepackage{hyperref}
\date{}

\title{A Perception Centred Self-driving System without HD Maps}

\author{%
Alan Sun\\
{\tt sun.qiao@wustl.edu}
}

\begin{document}
\maketitle

\begin{center}\textbf{Abstract}\\~\\\parbox{0.475\textwidth}{\em

Building a fully autonomous self-driving system has been discussed for more than 20 years yet remains unsolved. Previous systems have limited ability to scale. Their localization subsystem needs labor-intensive map recording for running in a new area, and the accuracy decreases after the changes occur in the environment. In this paper, a new localization method is proposed to solve the scalability problems, with a new method for detecting and making sense of diverse traffic lines. Like the way human drives, a self-driving system should not rely on an exact position to travel in most scenarios. As a result, without HD Maps, GPS or IMU, the proposed localization subsystem relies only on detecting driving-related features around (like lane lines, stop lines, and merging lane lines). For spotting and reasoning all these features, a new line detector is proposed and tested against multiple datasets. 

}\end{center}

\section{Introduction}

Ziegler's system \cite{6803933} can drive full-autonomously over 100 kilometers without any interruptions in 2014. Despite these early achievements, the industry leaders are still struggling to pass the necessary tests according to \cite{AVDR}. Obviously, we need to inspect why the current self-driving system is hard to implement and widely used. Current systems rely on HD Maps to produce centimeter-level accuracy of position. Readers are referred to \cite{8306879} for more about typical system architecture. The big question is that do we really need accurate positions?

Human drivers make driving decisions based on what they see. They make sense of the environment around and decide when to turn or keep the current driving direction. They cannot mark the exact position of themselves on a map, but they know how to travel through a complicated intersection based on the knowledge of which way they should take. Likewise, can we build a self-driving system without accurate locations?

In this paper, a new perception centered self-driving system is proposed and discussed in two driving scenarios: the cruising scenario and the turning scenario. The cruising scenario is when the vehicle cruises on parallel lanes. The turning scenario is when the vehicle drives through free spaces (defined as the drivable area outside of lanes, like intersections or parking area). 

The proposed system comes with several advantages in these two scenarios. Firstly, it does not rely on HD Maps. So it is easy to scale without recording new HD Maps. Secondly, the proposed feature detection method is not based on any specialized end-to-end deep learning solutions. Hence it is easy to debug and visualize. Also, it does not need additional time-consuming training process for scaling. Lastly, it performs more robustly with a severely changed environment (like seasons, weather or lighting condition). 

\begin{figure*}[!t]
\begin{center}
\includegraphics[width=\textwidth]{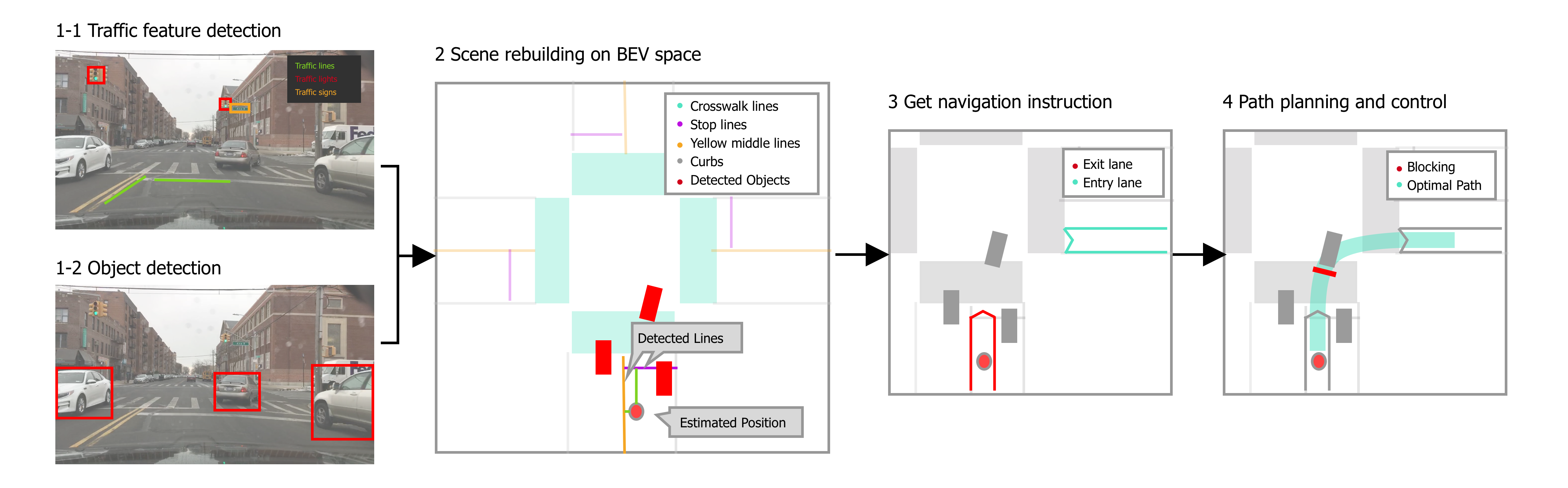}
\end{center}
   \caption{The workflow of the proposed perception centered system}
\label{fig: work flow}
\end{figure*}

Just like the human drivers, the system only involves with related visual features (defined as traffic features, including traffic lines, traffic lights and traffic signs). The workflow of the detection and localization subsystem is shown in Figure \ref{fig: work flow}. In the cruising scenario, we only need to finish the first step, including 1.1 and 1.2. In the turning scenario, we need to finish all four steps. Note that the vehicle position from the localization subsystem is based on the rebuilt scene rather than a global map. The localization subsystem also projects the rebuilt scene onto a digital map (like Google Map) to provide navigation instructions while crossing free spaces. The navigation instruction leads the car to travel from one exit to the target entrance of the free space. The path planning system and control system also works on the rebuilt scene. Hence they are map unrelated.

The proposed system relies on traffic lines (including curbs) for tracking the vehicle's position. Hence, the lines detector is the priority. We need a general lines detector for understanding complicated traffic lines on the road. The experiment covers several types of lines, including lane lines, stop lines, curbs, merging and splitting lines and intersections in a roundabout. For the popular lane lines detection problem, the proposed new traffic lines detector performs as good as other deep neural network supported approaches leveraging the prior knowledge of lines position and angles with easy erosion and clustering. This robust and straightforward method is then generalized and successfully detected other kinds of lines as well. After that, the process of localizing the position in the rebuilt scene will be discussed with examples and limitations. In that example, the system requires neither GPS signals nor IMU signals nor 3D HD Maps to locate the vehicle.

\section{Related Work}

What is a perception centred self-driving system? Most self-driving systems are relying on a map-based localization subsystem. They are categorized as localization centred systems because all other subsystems are working under the map space from the localization subsystem. The perception centred system uses a local scene, instead of a global map, as the working space for all other subsystems. Limited research have been done on this direction. One of the exceptions is \cite{DBLP:journals/corr/BojarskiTDFFGJM16} by Bojarski from Nvidia. In this work, they tried to build an end-to-end system from camera images to control signals with the help of augmented learning. It is also map-unrelated. However, this system only works for minimal lane-keeping tasks in the cruising scenario. It is not compatible to work with other subsystems, and the scalability is not tested for more sophisticated roads or sensor settings.

\subsection{Localization}

For most localization centred systems, all decision making and path planning are based on a centimetre level localization accuracy from their localization subsystem. Using GPS, with the aid of IMU, is a popular solution and provides accuracy better than 20 centimetres with SLAM over an HD Map \cite{8025618}. The problem of GPS is that the signals are not always available, and the result tends to drift accidentally. For quite a long time, SLAM is considered as the key to solving the localization problem for self-driving cars. The SLAM algorithm uses visual features stored in the HD Map to match features extracted from the live camera on the self-driving cars. Visual features are usually organized as bags of features (BoF) in the descriptor space. Without HD Maps or IMU, researchers can hardly reach the centimetres level accuracy like \cite{5476151} and \cite{7313496}. 

However, two problems of the SLAM based localization approach are tricky to solve. Firstly, the performance decreases once the environment changes. Light angle changes might cause different shadow shapes and season changes cause massive appearance changes on the trees and grass. These changes yield new visual features which cannot be matched with the recorded ones on the HD Map. This problem requires routine labour-intensive map recording once after the changes occur. Secondly, the localization result tends to drift after a long-range driving, and the error will accumulate with growing driven distance, as discussed in \cite{8025618}. The intrinsic reason of these problems is that the original SLAM algorithm is designed for indoor localization problems where dramatic environment changes or long-distance moving is not considered. Hence these problems are hard to eliminate.

Recent researchers, like Ma \cite{Ma_2019}, started to use as less visual features as possible for localization. Besides saving the storage for the BoF of these features, using fewer features decrease the risk of being affected by the environment changes \cite{Toft_2018_ECCV}. 

This trend brings the idea of using minimal features for localization. The LaneLoc system proposed by Schreiber \cite{6629509} tried to use the exact appearance of lane markings for matching from pre-recorded maps. This approach could be seen as counting the number of dashed fragments the vehicle travelled to localize the car itself. This approach still has several limitations. Firstly, it will not work on a solid line situation and ends up with only relying on IMU without any visual aids. Secondly, the exact appearance will eventually change one day in the future. Think about the time when those dashed lines were repainted or worn out, which are both prevalent cases. Thirdly, the performance is very fragile. Slight turbulence, like occlusions or heavy shadows, will make the system omit one or more fragments and yield a steady error as a result. Lastly, the labelling process is both complicated and hard to finish accurately, as discussed by Schreiber in their paper. Our proposed system solved these limitations by abstracting line features further to types and directions by the proposed lines detector.

\subsection{Traffic Line Detection}

The traffic line detection, or the lane detection which is a narrower problem, was the essence of many early driving assistant systems \cite{10.1007/s00138-018-0977-0} like Lane Departure Warning System (LDWS) and Lane Keeping Assist System (LKAS). Many researchers, like Kim \cite{inproceedings}, used Convolutional Neural Network (CNN) to reduce noise and get the segmentation of the markings of those lines. Wang \cite{Wang_2020} used shape extracted from OpenStreetMap (OSM) as prior knowledge to help detect the lanes. Some problems remain for the CNN supported approaches. 

Firstly, they still can not solve the long-tail challenging situations because CNNs heavily relies on the distribution of the training dataset. As a result, CNN generally works terribly in rare situations. Secondly, the segmentation result of the CNN approaches often cause blurry edges when it is not confident about the prediction. These blurry edges come with difficulty for the following algorithms when they try to form a line from these ambiguous pixels. Lastly, CNNs are significantly dataset related. They tend to work well only on the dataset they have been trained on \cite{4938723}. This limitation is because that different datasets and sensor settings tend to create distinctive patterns of noise in the images. For example, in the KITTI dataset \cite{6248074}, the same line marks show different appearances in different locations under the BEV space. Lines far from the camera shows clear artifacts caused by the BEV transformation. The self-driving related datasets are often covering just one type of the available camera settings. A vast and comprehensive dataset like MS-COCO \cite{Lin_2014} for the object detection task does not exist for now.

As a result, CNN was not used for lines detection in this paper. The proposed lines detector leverages the lines information from a topology map, similar to what Wang did in \cite{Wang_2020} from the OSM, as prior knowledge to help. The proposed lines detector separates different line types to boost the performance even more by using different lines detector for each type of lines (solid or dashed lines, straight or curved lines). It also used a sliding window to detect and connect traffic lines, similar to what Tsai did in \cite{4621271}. The sliding window approach is proved to be both robust and easy to visualize for debugging.

\section{System Design}

The overall workflow is shown in Figure \ref{fig: work flow}. In the cruising scenario, the detection subsystem will finish the part 1.1 and 1.2 to give the current lane number of the vehicle, and that is enough for generating a driving path and control signal without involving the localization system at all. However, the detection system needs to continuously detect the traffic features for the next traffic part (could be another lane ahead or a free space connected with an exit). The order of the series of traffic features are based on the topology map. 

\begin{figure}
 \center
  \includegraphics[width=\columnwidth]{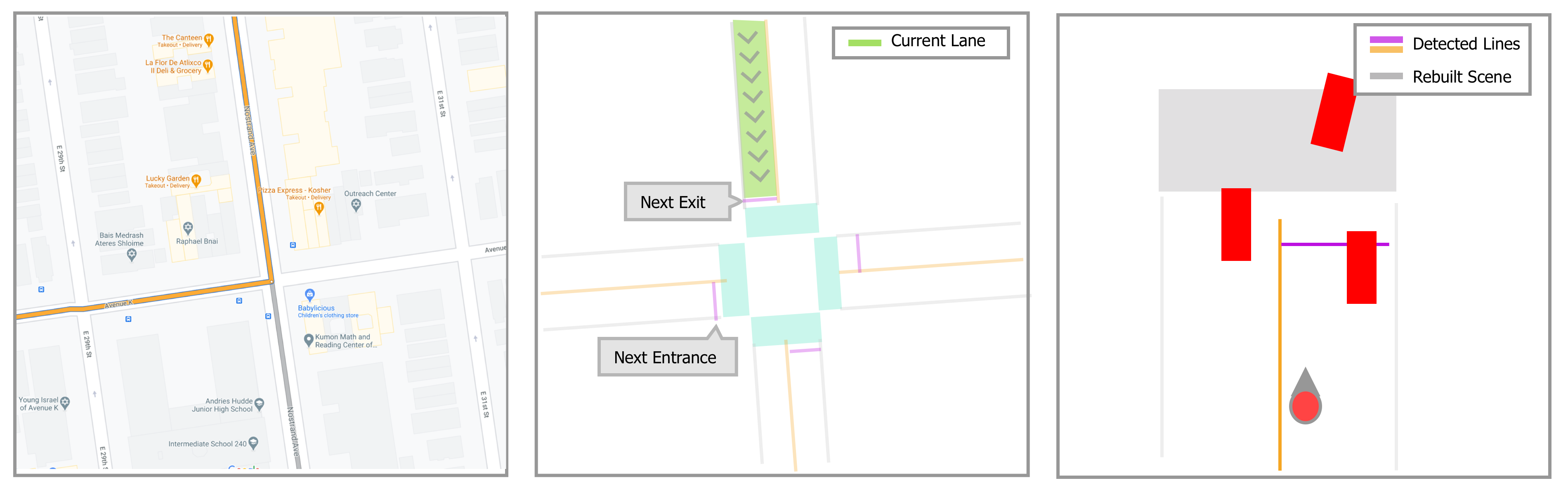}
  \caption{The left is a digital map used for navigation, the middle is the topology map, the right is the rebuilt scene}
  \label{fig: maps}
\end{figure}

The topological map, being used as the descriptor space for matching with the digital map and the rebuilt scene, is the centre and the relationship is shown in Figure \ref{fig: maps}. The topology map should be drawn before the system can run on a new area. The topology map also provides lane information helping lines detection as prior knowledge and helps the vehicle to change to a preferred lane in advance. The topology map contains the following information: 

\begin{itemize}
\item Lanes information: (1) the lines information on both sides (like straight yellow lines on the left and straight curb on the right), (2) ending information (like ends with a stop line or merges with other lanes on the left),  (3) direction information (like starting direction, turning angle limitation for each window), (4) neighbour lanes used for lane changing while cruising, (5) connected entrance and exit numbers, (6) traffic rules metadata (like speed limits), (7) status (like normal, under maintenance or closed under specific time windows)
\item Entrance and exit: (1) position, (2) direction, (3) the relationship (an N to N relationship) with each other.
\item Free spaces: (1) detectable traffic features used for localization (including stop lines, crosswalk lines, traffic lights, traffic signs, lines of adjacent lanes) and their relative position in a real-world scale, (2) adjacent entrance and exit numbers, (3) traffic rules (like speed limits), (4) status
\end{itemize}

\subsection{Matching digital map with topology map for navigation}

Each turning point on the digital map is used for finding a nearest entrance-exit pair which have the correlated directions. Define $\boldsymbol{T} = \left \{(\lambda_t, \phi_t), \alpha_t, \beta_t \right \}$  as the set of all turning points on the digital map, where $\lambda_t$ and $\varphi_t$ is the latitude and longitude of turning point $t$, $\alpha_t$ is the direction before the turning and $\beta_t$ is the direction after the turning. $\boldsymbol{D}_{in} = \left \{(\lambda_d, \phi_d), \beta_d\right \}$ and $\boldsymbol{D}_{out} = \left \{(\lambda_d, \phi_d), \alpha_d\right \}$ are the set of all entry points and all exit points. The score function f is the multiplication of g and h, as equation \ref{eq:1}, where g is the Euclidean distance between two points and h is the difference of two angles, defined as $g=\left \| d_{in}, t \right \|+\left \| d_{out}, t \right \|$ and $h=\left | \alpha_t, \alpha_{in}  \right | + \left | \beta_t, \beta_{out}  \right |$. The $\boldsymbol{P} = \left \{ (d_{in}, d_{out}) \right \}$ is the set of all legal pairs of entrance and exits. All legal pairs should connect with a same free space and follow the traffic law. For example, the exit on the end of a right turning lane cannot pair with the entrance ahead with the same direction. The optimal pair for a minimal f score is the matched result with the condition of $(d_{in}^*, d_{out}^*) \in \boldsymbol{P}$. This method assumes the turning point on the digital map is the center point of the target exit and the target entrance.

\begin{align}
    f(d_{in}, d_{out}, t) = g(d_{in}, d_{out}, t)*h(d_{in}, d_{out}, t)
\label{eq:1}
\end{align}

The data of $\boldsymbol{P}$ and $\boldsymbol{D}$ are manually initialized as part of the topology map. These data usually do not need to be changed unless the traffic features are changed. For example, an intersection was updated with an additional right changing lane or new construction on the road updated the lane changing rules temporarily. The maintenance of the topology map is easy, fast since we only need to change the lane data in the sets of $\boldsymbol{P}$  and $\boldsymbol{D}$.

\subsection{Matching topology map with perception scene for localization}

\begin{figure}
 \center
  \includegraphics[width=\columnwidth]{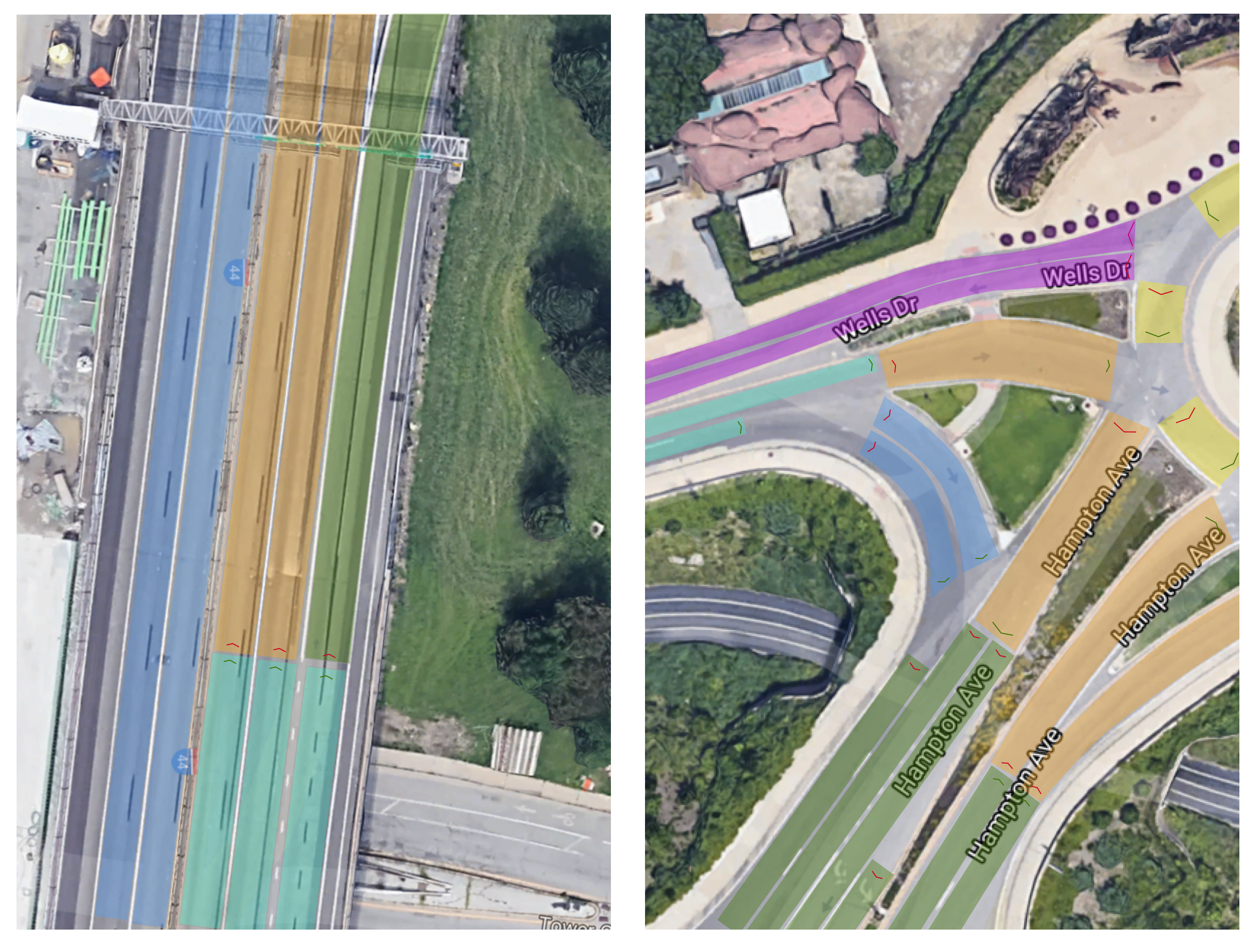}
  \caption{An illustration of lane sets under two different situations overlapping on a satellite map, the left is on highway exit, the right is a complicated lane topology near a roundabout, the red and green arrows represent entrances and exits of those lanes}
  \label{fig: lane_sets}
\end{figure}

Lanes form two kinds of lane sets: driving lane sets and detectable lane sets. The driving lane sets provide information about lane changing behaviour and traffic laws, like speed limits. Two examples of driving lane sets are illustrated in Figure \ref{fig: lane_sets}. The vehicle can change to other lanes within the same driving lane set. The target lane and original lane information will be passed towards other following subsystems to act and finish the changing maneuver while lane changing.

The detectable lane sets provide information about how to detect these lanes. Lanes with either the same travelling direction or opposite ones can be grouped into the same set. Each detectable lane sets has left and right line types, lane width (used as detection aid, but restrictions), dashed line intervals, suggested detection window size and other metadata which can be added at one's convenience. A detectable lane set must have at least two sides of lines information used for tracking lanes. The lane width follows the priority of (1) the width between two detected lines, (2) the width of other detected lanes within the same detectable lane sets, (3) equally divided width if two lines (probably are curbs) of the whole set are detected, (4) the default lane width of the detectable lane sets.  For example:

\begin{itemize}
\item When there are two lanes travelling in opposite directions, and there is no middle line to separate these two lanes, if both sides are detected, the space in between will be divided by two for the width of each lane. If the vehicle only detected the right side (assume under right-hand driving condition), the lane width for the current lane is the default lane width of the lane set.
\item When there are four lanes travelling in opposite directions by two groups of two lanes, the middle line is a solid line, and the line between two same direction lanes is a dashed line. The number one lane (counting from the right) is the space between the curb and the dashed line, and the number two lane is the space between the middle solid line and the dashed line. If the vehicle cannot detect the curb to get the lane width of lane number one, the width of number two will be used for the width of lane number one.
\end{itemize}

For the cruising driving scenario, we care about two questions: (1) which lane set we are in (to prepare for the next exit) and (2) which is the ego lane from the lane set. For these two questions, the system relies on either initializing the lane number at the beginning of the currently running period or initializing the lane number after driving through a free space through a specific entrance. The detection system verifies and corrects the current lane set and lane number by matching detected types of lane lines with the ones from the topology map. The detection system provides four line detectors for each type of lines: (1) solid straight line, (2) solid curve, (3) dashed straight line, (4) dashed curve. In the remainder of this paper, curbs are considered as the same as traffic lines without further clearance.

The changes of the types of lane lines usually represent an end of the current lane. If is possible that there will be multiple types of lines in one side of a lane, the system uses the detector for the highest level type, because they are more complicated and can handle the task of detecting low-level types. These levels (one is the highest level and four is the lowest) are: (1) Dashed curves, (2) Solid curves, (3) Dashed straight lines, (4) Solid straight lines.

\begin{figure}
 \center
  \includegraphics[width=\columnwidth]{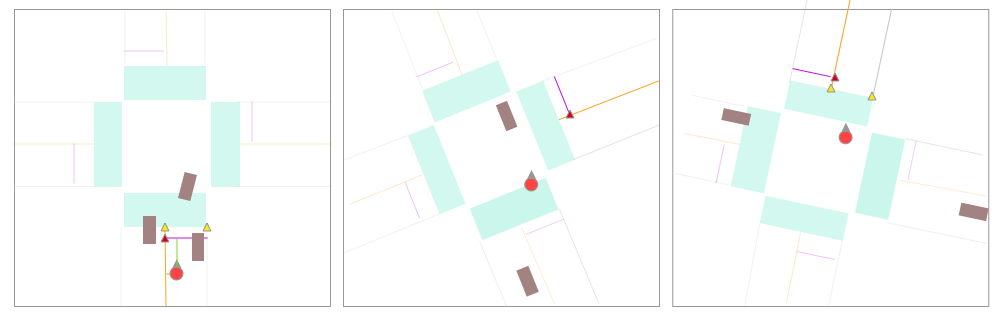}
  \caption{How the vehicle locate itself using lines detection results. The yellow triangle is a weak anchor and the red triangle is a strong anchor.}
  \label{fig: turning_loc}
\end{figure}

For the turning scenario, the detection subsystem only needs to detect one pair of non-parallel lines to form an anchor to rebuild the scene. For example, under the intersection scenario shown in Figure \ref{fig: turning_loc}, the middle lane line and the stop line are enough for a strong anchor to rebuild the scene based on the given relative position from the topology map. The target entrance on the right side can be predicted and used for path planning. Once the vehicle has driven into the free space passing the stop line which will no longer be detected, the stop line of the target lane will be detected and provide a strong anchor to follow up. The starting point of the target lane will form a weak anchor as additional clues for localization. 

The detection of anchors might be effected by occlusions caused by other objects on the road. In other situations, there is a chance when the vehicle is crossing a large intersection, the vehicle will have no available anchor in sight in some area. The target lane direction and the current drivable area, as a backup, will aid the vehicle to finish the turning. The free space situation ends with positive detection of the next detectable lane set. If there are multiple lines parallel with each other nearby, the system assumes the detected one is the nearest one based on the current lane level position. 

The system needs to be initialized at the beginning of each run based on GPS signals and the current driving direction from the gyroscope to tell the system which lane the vehicle is on. The GPS signal does not need to be centimetre-level accurate, and the detection subsystem will update the lane number, relying on counting the line numbers between the vehicle and the detected curbs.

This paper does not cover behaviour decision among crossing lanes because this can be considered as a separated and solved problem thanks to previous research like \cite{6803933}. This behaviour decision includes behaviours, like yielding to vehicles coming out from other merging lanes. These rules are universal and consistent.


 
\subsection{General lines detector}
\label{sec:lines detection}

\begin{figure}
 \center
  \includegraphics[width=\columnwidth]{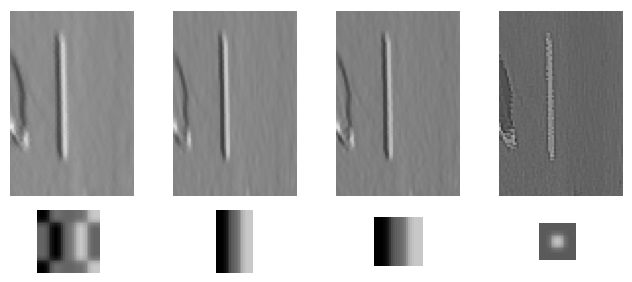}
  \caption{Four results for different convolution kernels. The first one is the proposed one and the last one is a typical square edge detection kernel. The result on the left is more smooth and cleaner than the ones on the right in the noisy area.}
  \label{fig: kernels}
\end{figure}

\begin{figure}
 \center
  \includegraphics[width=\columnwidth]{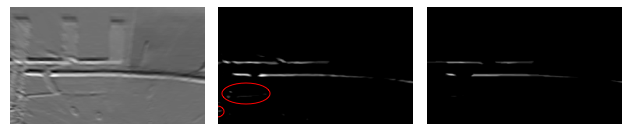}
  \caption{Directional erosion eliminates strong noise in the red circles while detecting stop lines.}
  \label{fig: erosion}
\end{figure}

\begin{figure}
 \center
  \includegraphics[scale=0.4]{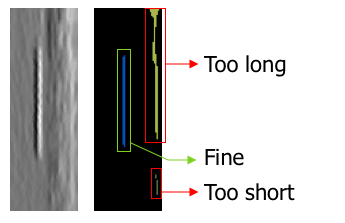}
  \caption{An example of how length information helps to filter noise within a single window.}
  \label{fig: length}
\end{figure}

The proposed lines detector in the detection subsystem can detect diverse types of lines. The code for lane lines detection for the KITTI dataset can be found on this \href{https://github.com/larksq/lane_detector_for_KITTI}{repository}. These types of lines were tested: (1) lane lines, (2) curbs, (3) stop lines, (4) merging or splitting points of two lines (pair of lines), (5)  special lane lines or curbs (which are not parallel to the current ego lane). The lines detection problem was dissected by tracing back to the most significant visual feature of the lines, which is their long and narrow appearance. A sliding window was used to follow possible lines. All noise without this narrow feature was eliminated by applying these methods:

\begin{itemize}
\item Region Restriction: The detection subsystem leverage a given prior knowledge about the starting points to eliminate noise in unrelated regions. This knowledge comes from either previous lines detection results or predicted by the positive detection results of neighbour lines with given lane width from the topology map. For dashed lines, the sliding window moves at a step size of dash segment intervals given from the topology map to make sure optimal detecting position for each segment. The system tolerates minor errors for this interval distance. The more knowledge we know about the lines, the smaller window for detection we can use. A smaller region of interest gives better resilience for challenges, helps the segment normalize better and speeds up the lines detection process.
\item Special Convolution Kernel: The system uses a special kernel, as shown in Figure \ref{fig: kernels}. This proposed kernel helps to produce a cleaner result in the Hough space for the next steps with less noise. Also, this kernel is more friendly for detecting curves, merging lines and splitting lines than the simple vertical kernel.
\item Directional Erosion: The system uses a special directional erosion structuring element to erode noise which is not spanning through a specific direction ($A \ominus B$, $A$ is the pixels in the window and $B$ is a 5 by 1 narrow structuring element), as illustrated in Figure \ref{fig: erosion}. The direction of the target lines is given from the topology map. In a sliding window, the line segment can be considered as a straight line. Sharp turning lines or circles will also be eroded into small segments which will be filtered out. Though there are some other more complicated ways to leverage the information of direction for lines detection \cite{Finding_multiple_lanes_in_urban_road_networks_with_vision_and_LIDAR}, the directional erosion is the simplest and it works.
\item Types of Lines: The system leverages prior knowledge of the types of the lines to get a better performance. For curves in each detection window, the turning angles are restricted to the thresholds, which is usually very small given from the topology map. For straight lines, we can use a much narrower window for detection. For dashed lines, the marks which are too long or too short will be filtered out, as shown in Figure \ref{fig: length}. The topology map gives the length of segments of the dashed lines.
\end{itemize}

The proposed lines detector uses the Y channel from the YUV color channels since it was proved to perform better by Lin in \cite{5489518}. The system works on the Bird-Eye-View (BEV) space since we can leverage the prior knowledge of those lines without predicting the camera pose or estimating the vanishing point (VP) \cite{8303759}. More about the homography transformation from the camera image to a BEV space with a given camera pose can be found in \cite{4459093}.

\begin{figure}
 \center
  \includegraphics[scale=0.4]{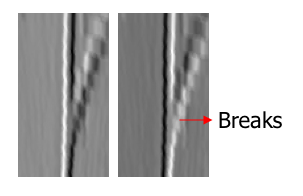}
  \caption{Illustrate when lines splits, left side and right side line detection result will not agree with each other}
  \label{fig: split}
\end{figure}

For the feature detection on the Hough space, a low-high-low kernel was widely used by \cite{650851}, \cite{10.1093/ietisy/e89-d.7.2092} and \cite{4538013}. I changed it to a low-middle-high kernel and then mirror it to make the detection on the left and right side separately. So we can detect merging and splitting points and their directions (merging from / splitting to the left or the right) by comparing the lengths of these two lines detection results. For example, at the place a line is splitting to the right, the line detection from the right side will break coming with a shorter length of the line than the left side, as shown in Figure \ref{fig: split}. To separate splitting and merging, two additional windows will be created facing upwards and downwards. Positive result of lines in the upwards window means splitting and positive result in the downwards window means merging.

Lastly, the procedure for stop lines detection is as follows. After the detection of a window, if the line is broken in the upper end, two side windows will be created. A horizontal line detection, using horizontal convolution kernel and erosion structure, will be applied to detect the stop lines. If the result is positive, then this lane line is marked as finished, and no window will be created above. 

The overall process of lines detector is shown as pseudo code in algorithm \ref{algorithm: normal}. For special lines which are not parallel to the current ego lines, the problem is we do not have an initial position for the sliding window to start. However, we can still use the direction information from the topology map. Spotting the anchors from the target entrance while turning in free spaces is one of the situations which requires detecting special lines, as shown as in Figure \ref{fig: turning_loc}. The algorithm is a little different, shown as pseudo code in algorithm \ref{algorithm: special}. The image after erosion is cut into several blocks. The blocks containing valid pixels are used for forming windows.


\begin{algorithm}
 \SetAlgoLined
 \KwData{Input Image to detect and topology map information}
 \KwResult{Detected lines}
 Initialize the first window\;
 \While{sliding window does not go out of the image}{
  Cut pixels in the window\;
  Rotate the window\;
  Commit Convolution\;
  Commit Directional Erosion\;
  Cluster pixels\;
  Form lines for all candidates\;
  Filter and get the result for current window\;
 }
 Connect results and form a line\;
 \caption{The lines detector}
 \label{algorithm: normal}
\end{algorithm}


\begin{algorithm}
 \KwData{Input Image to detect and topology map information}
 \KwResult{Detected lines}
 Rotate the image\;
 Commit Convolution\;
 Commit Directional Erosion\;
 Get valid pixel blocks\;
 Form valid blocks into windows\;
 Cut pixels for each window\;
 \For{each window}{
  Cluster pixels\;
  Form lines for all candidates\;
  Filter and get the result for current window\;
 }
 Connect and merge similar lines\;
  \eIf{Any positive result in any window}{
   Return the longest detection result\;
   }{
   Return negative as a result\;
  }
 \caption{The special lines detector}
 \label{algorithm: special}
\end{algorithm}

\section{Results}

The earlier part of this chapter shows the proposed general lines detector is robust to typical noise on the road, works well under different lighting conditions and detect multiple types of lines. The later part of this chapter shows how the localization method helps the vehicle travels through an intersection in the turning scenario. 

For lane lines detection, the method was tested on KITTI \cite{6248074} and Cityscapes \cite{DBLP:journals/corr/CordtsORREBFRS16}. For general traffic lines detection, The proposed method was tested on the Berkeley deep drive (BDD 100k) \cite{DBLP:journals/corr/abs-1805-04687}, KITTI and a self-recorded video. These results of general lines detection cannot be compared to other methods due to lacking metrics. At last, the BDD 100k dataset and images from a self-recorded video are used for testing the localization method while passing free spaces.

\subsection{Lane lines detection}

\begin{figure}
 \center
  \includegraphics[width=\columnwidth]{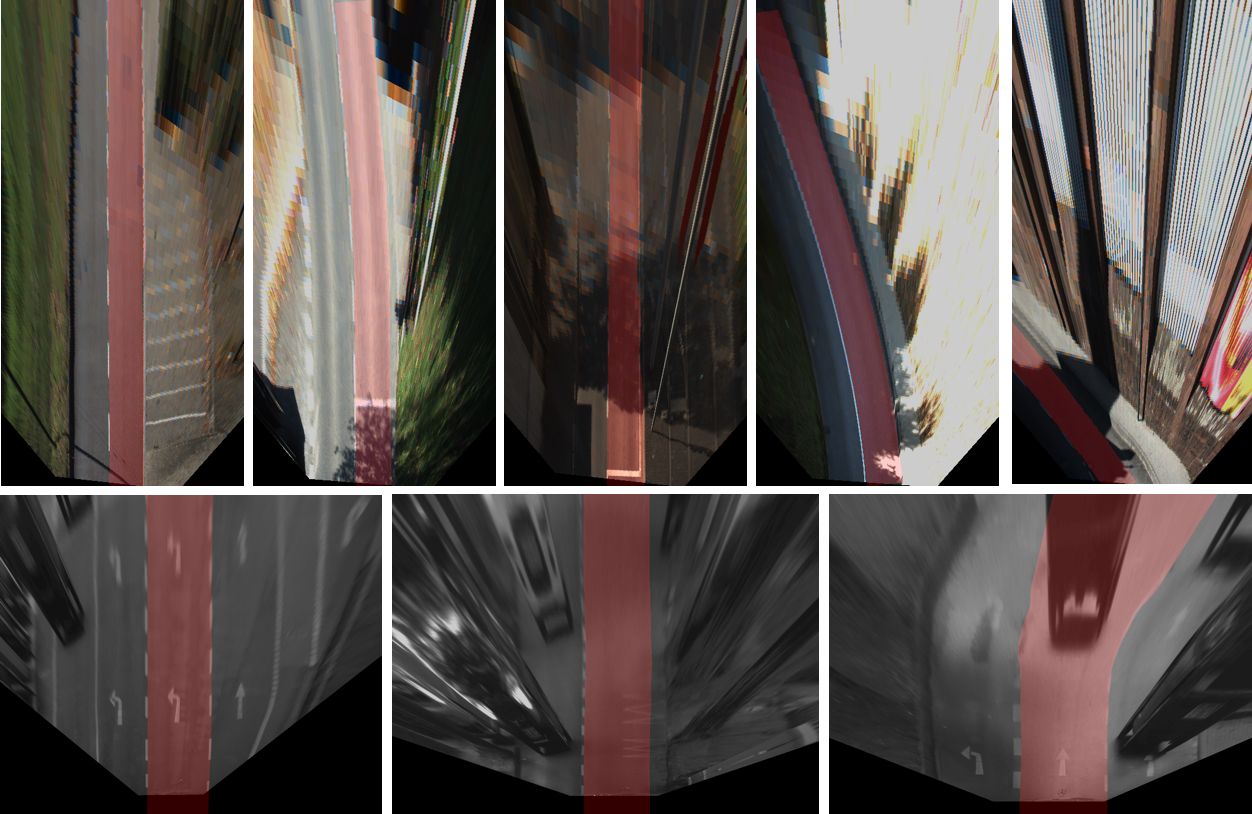}
  \caption{The first row is some of the detection results of KITTI-UM. The second row is some of the detection results of Cityscapes. The red is the converted ego lane based on lines detection.}
  \label{fig: KITTI_lines}
\end{figure}

The proposed lines detector, ECPrior (Erosion and Cluster with prior knowledge), perform as good as other deep neural network supporting approaches \cite{oliveira2016eifficient} \cite{wang2020map} \cite{chen2017rbnet} \cite{lyu2019road} based on the KITTI behaviour evaluation \cite{6728473} metric. The result is shown in table \ref{table: result 1}. Some of the detection results are shown in the first row of Figure \ref{fig: KITTI_lines}. The proposed detector does not include object detection; hence it will be affected by other cars close to the lines. A typical object detector can be added before to get a better result, like Satzoda did in \cite{6910058}. The object detection is usually a separate module, and the same feature should not be implemented again in the lines detection module. The proposed lines detector works equally fine on Cityscape showing its scalability, as shown in the second row of Figure \ref{fig: KITTI_lines}, despite they have very different object aspect ratio from the aspect ratio of images from KITTI.

\begin{table}[h!]
\centering
\begin{tabular}{c | c | c | c}
{\bf Method} & {\bf HR-30} & {\bf PRE-40} & {\bf F1-40}\\ \hline
CyberMELD  & 97.55 \% & 94.57 \% & 89.66 \% \\
RBNet & 95.92 \% & 95.56 \% & 87.21 \% \\
RoadNet3 & 95.57 \% & 94.57 \% & 83.72 \% \\
ECPrior (Mine) & 93.96 \% & 96.70 \% & 91.86 \% \\
Up-Conv-Poly & 93.14 \% & 90.11 \% & 83.72 \% 
\end{tabular}
\caption{KITTI (UM Lane) lane lines detection result}
\label{table: result 1}
\end{table}

\begin{figure}
 \center
  \includegraphics[width=\columnwidth]{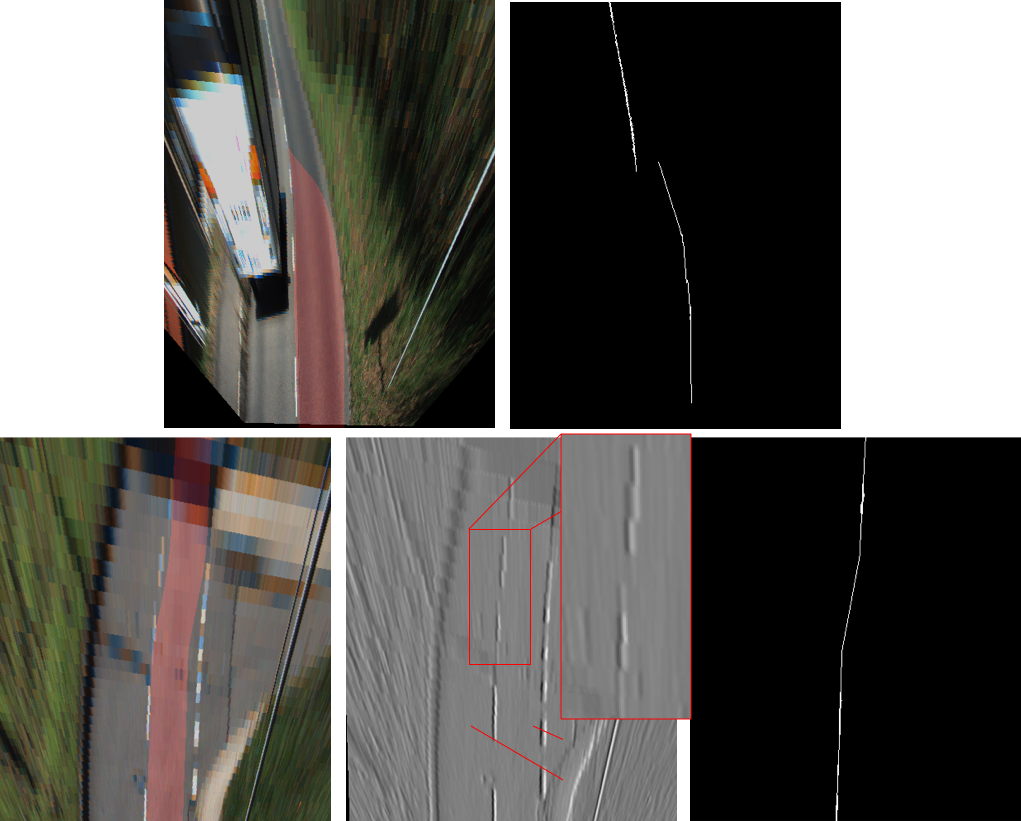}
  \caption{Two fail cases in each row: (1) the first image is the ego lane result and the second is the lines detection result of the right side curb; (2) the middle image is the gradience showing clear manifests and the lines detection results shown in the right image.}
  \label{fig: lines_neg}
\end{figure}

The limitations of ECPrior are: 

\begin{itemize}
\item Like all other methods, ECPrior relies on a stable and accurate BEV transformation. The transformation is hard to be accurate when the ground is not flat. Although the deep neural networks can learn to avoid this for a specific dataset, it is still hard to scale over different datasets. When it comes to non-flat surfaces, the width of a lane might shrink, as shown in the first fail case in Figure \ref{fig: lines_neg}. Dynamic adjustment of the window width can avoid windows from merging. ECPrior can tolerate minor distortion of the BEV transformation.
\item Because ECPrior is for general cases, the input images should not have special manifests which would disturb the detector, as shown in the second fail case in Figure \ref{fig: lines_neg}. For KITTI, these manifests are mainly caused by the BEV transformation over low quality areas.
\end{itemize}

\subsection{General lines detection}

\begin{figure}
 \center
  \includegraphics[width=\columnwidth]{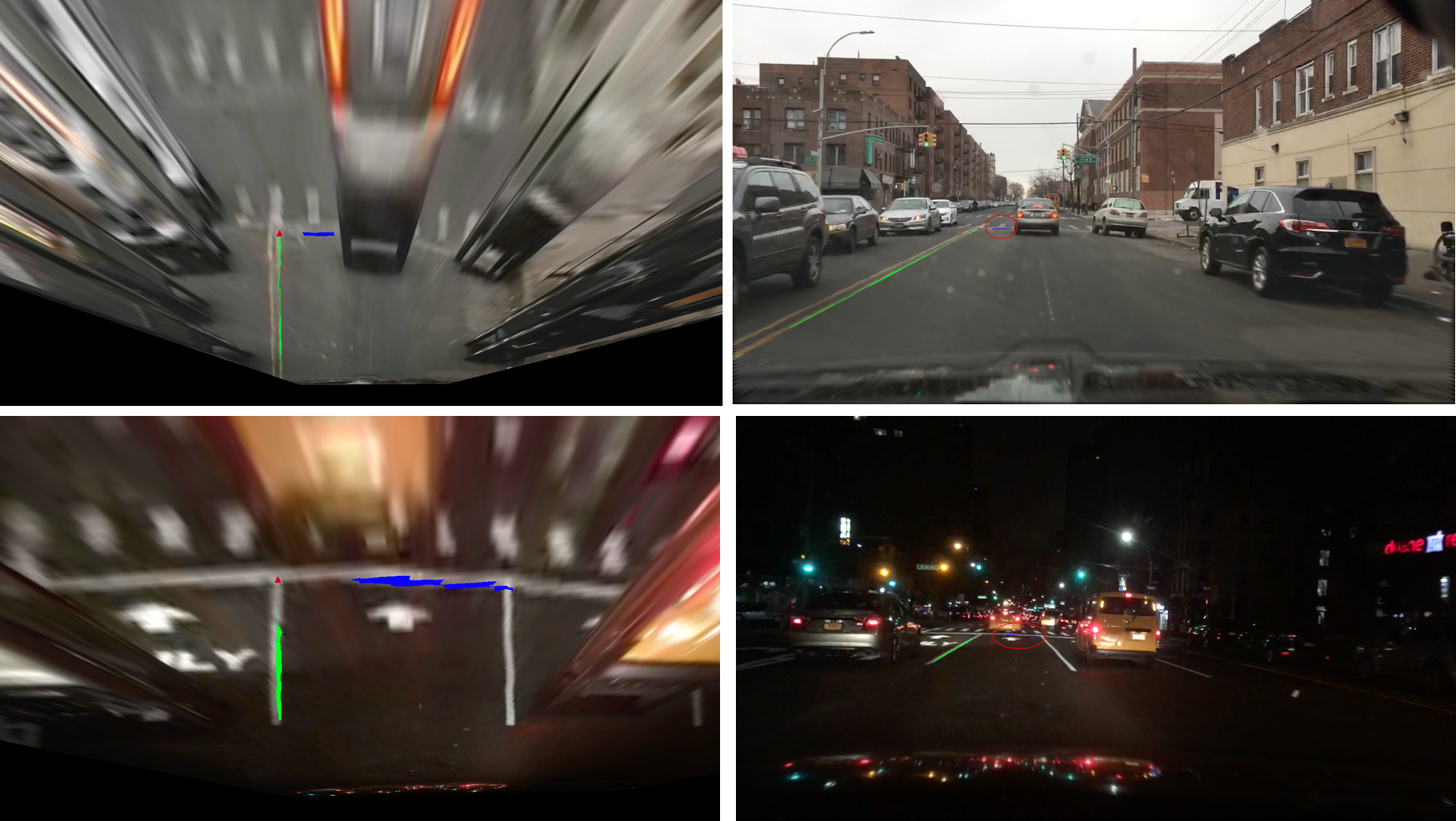}
  \caption{Stop lines detected results on BDD 100K dataset. Blue pixels are the detected stop line and green pixels are the guiding lane line of that stop line.}
  \label{fig: bdd100k_stoplines}
\end{figure}

ECPrior can solve the problem caused by shadows or short breaks for general lines detection. ECPrior is also proved to be robust with different lighting conditions. For stop lines, images from the BDD 100k was used for testing. The result is shown in \ref{fig: bdd100k_stoplines}. The upper case in that image is under a lightly snowing daylight environment, and the lower case in that image is in a night lighting environment. In both cases, ECPrior successfully detects the stop lines ahead. 

\begin{figure}
 \center
  \includegraphics[width=\columnwidth]{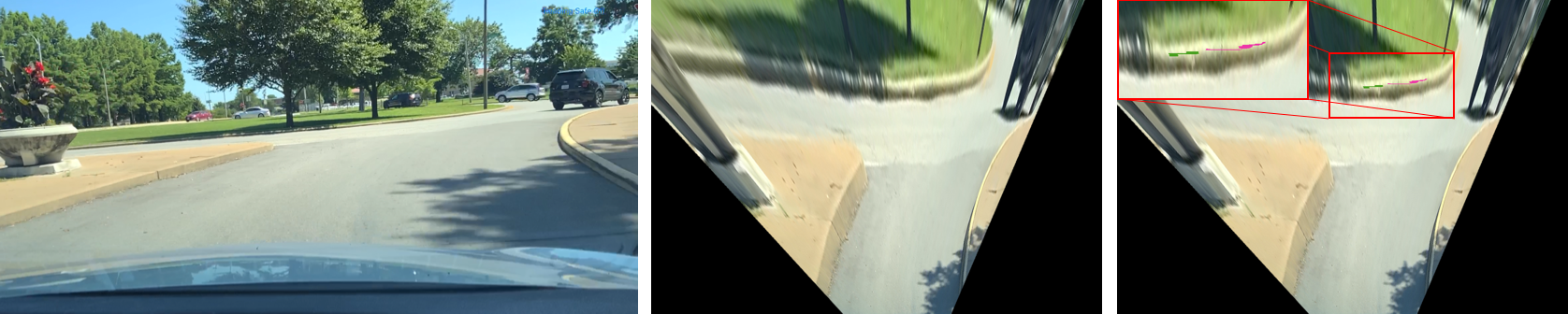}
  \caption{Detecting the inner side of the roundabout is an example of detecting a special line with only its direction given. The top 2 results are shown as red and blue in the last image in the red box.}
  \label{fig: roundabout}
\end{figure}

ECPrior also detects special lines well. A self-recorded video was used for testing. An example in Figure \ref{fig: roundabout} shows the ability to detect special lines under a turning scenario travelling into a roundabout. In this situation, ECPrior needs to detect the rear inner side of the roundabout. The left side curb of the current lane and the inner side curb of the roundabout can then form a strong anchor used to rebuild the scene of the free space for localization. 

ECPrior uses intense erosion and threshold so that only a small portion of target lines will be detected at the pixel level. Hence the ECPrior detector is not a pixel-level detector. ECPrior, as an intact line detection module, provides lines detection result using regression for dash line segments and straight lines and using Spline for the others. ECPrior inevitably relies on an accurate BEV transformation to leverage the prior knowledge of the lines. Distortion due to camera behind the windshield or problematic camera settings also cause a narrower efficient area for general lines detection, at that situation only lines lie in the middle of the front can be detected. As an example, the detector failed to detect the left side of the inner curb due to distortion in Figure \ref{fig: roundabout}.

\subsection{Localization}

Based on these results from previous examples, strong and weak anchors can be established to locate the vehicle in the turning scenario. The proposed localization approach relies on neither GPS nor IMU for vehicles to travel through urban areas. The system provides a stable and accurate position based on the rebuilt scene for path planning and control subsystems in the turning scenario. For the cruising scenario, the detection system gives a lane level localization result (which lane the vehicle is on) which is enough for the following subsystems.

There are several limitations for using the naive approach of my proposed system for localization. Firstly, the proposed localization method relies on visual clues of specific traffic features. Heavy occlusion blocking most of the target traffic lines will affect the location result in some degree. In one situation, the vehicle was approaching the intersection with heavy traffic ahead, blocking most of the coming stop lines. The localization system did not spot anchors until when the vehicle was very close to the stop lines, producing a short reaction time to stop for the following subsystems. In another situation, the vehicle was about to turn right into a small allay based on the navigation. Several parking vehicles blocked the view of the right side curb. Hence the detection subsystem did not detect the right turning feature for the allay and make the vehicle miss the target turning. 

In the first situation, we can involve the behaviours of other vehicles as an input for the localization subsystem, like the way Gao leveraged the position of other vehicles in \cite{5204359}. For example, when we detect a line of stopping vehicles, we can assume the position of the first stopping car is indicating the position of the stop line to form a prediction to extend the reaction time for the following subsystems. In the second situation, a more comprehensive drivable area analysis will show a right side road extension indicating the allay. Additionally, the localization subsystem is compatible with traffic lights, traffic signs and GPS as pieces of additional information to help.

\section{Conclusions}

This paper proposes a new perception centered self-driving system and focuses on testing the proposed general lines detector, ECPrior, and the localization method on several urban cases. The proposed system design is a skeleton and a starting point with all potentials to work with additional modules to get better performance. For example, users can try to apply the method by Hillel in \cite{hillel2014recent} to get rid of the lens flare to make the detection of ECPrior more robust when driving towards the sunshine. The potential is much more promising than other deep neural networks based detection methods. And diverse types of scenes rebuilding can be discussed in future works. Places like indoor parking area without GPS signals will heavily rely on the rebuilt scene to localize the vehicle. Hence they should be prioritized. 

In the end, I appeal to the community to reconsider the necessity of using SIFT like visual features for localization, as well as the need for relying on deep neural networks for traffic lines detection in the context of self-driving.

{\small
\bibliographystyle{unsrt}
\bibliography{refs} 
}

\end{document}